\documentclass[11pt,a4paper]{article}
\usepackage[hyperref]{acl2017}
\usepackage{times}
\usepackage{latexsym}
\usepackage{url}

\usepackage{amsmath}
\usepackage{multirow}
\usepackage{url}
\usepackage{graphicx}
 \usepackage{subfigure}
\graphicspath{ {Figure/} }
\usepackage{amsmath, amsfonts}
\usepackage{algorithm}
\usepackage[noend]{algpseudocode}
\usepackage{multirow}

\usepackage{xcolor}

\aclfinalcopy

\title{Prepositions in Context}
\author{Hongyu Gong, Jiaqi Mu, Suma Bhat, Pramod Viswanath\\
hgong6@illinois.edu, jiaqimu2@illinois.edu, spbhat2@illinois.edu, pramodv@illinois.edu
\\
Department of Electrical and Computer Engineering\\ 
University of Illinois at Urbana Champaign, USA
}
\date{}

\begin{document}
\maketitle

\begin{abstract}
Prepositions are  highly polysemous, and their variegated senses encode significant semantic information. In this paper we match each preposition's complement and attachment  and their interplay crucially to the geometry of the word vectors to the left and right of the preposition. Extracting such features from the vast number of instances of each preposition and clustering them makes for an efficient preposition sense disambigution (PSD) algorithm, which is comparable to and better than state-of-the-art on two benchmark datasets. Our reliance on no external linguistic resource allows us to scale the PSD  algorithm  to a large WikiCorpus and learn sense-specific preposition representations -- which we show  to encode   semantic relations and  paraphrasing of verb particle compounds, via simple vector  operations. 
\end{abstract}

\section{Introduction}

%


  Prepositions form a closed class showing no inflectional variation and  are some of the most frequent words. 
  Yet,  they remain largely under-explored in computational linguistics owing to  their highly polysemous nature and frequent participation in idiomatic expressions \cite{saint2006syntax}. In this paper,  we study the problem of sense disambiguation for prepositions and the related problem of their distributed representation.



Computationally, two different views of prepositions are standard: sometimes  they are treated as being semantically vacuous and other times as being indiscriminate in association owing to their polysemy. Language processing tasks   operating at a surface  level of words treat them as  stop words and  disregard them (e.g., bag-of-words models), whereas those  harnessing the syntax and semantics of words resort to the latter. In doing so, the latter tackles the  challenges brought about by prepositional ambiguity in tasks such as prepositional phrase attachment \cite{ratnaparkhi1994maximum,collins1995prepositional}, semantic role labeling \cite{srikumar2013modeling} and downstream applications such as grammatical error correction \cite{chodorow2007detection} and machine translation \cite{shilon2012incorporating}.

The highly polysemous nature of prepositions drives several syntactic and semantic processes. For instance, the preposition \emph{with} has 18 senses listed in The Preposition Project (TPP) \cite{litkowski2005preposition}, examples of which, are shown in Table~\ref{tab:with_senses}. We notice that \emph{with} indicates an emotional state in \emph{with confusion} and refers to an accompanier in \emph{combine with}, while it suggests the idea of a tool or means in \emph{wash with water}. Thus,  preposition sense disambiguation (PSD) is vital for natural language understanding 
and a closer look at the function  of prepositions in specific contexts is an important computational step.


Antecedent approaches to PSD (for instance, \cite{ye2007melb,hovy2011models}) have relied on linguistic tools and resources (the minimum of which involves  dependency parsers and POS taggers) to capture the crucial contextual information of prepositions.  We depart by using no linguistic resources or tools other than a set of  word representations (trained on a  large  corpus). 
We interpret preposition senses as groups of similar contexts, where each instance of the preposition `sense' is  represented as  vectors of context-dependent features. We find that a simple feature extraction that  creatively harnesses the geometry of word representations   yields  a scalable algorithm that can reach near and even beat state-of-the-art performance on two  benchmark datasets (SemEval 2007 and OEC); this is true in both unsupervised and supervised PSD settings.



\begin{table}[htbp]
\centering
\resizebox{0.5\textwidth}{!}{
\begin{tabular}{|c|}
\hline
 Sentence (TPP sense) \\ \hline
 She blinked \textbf{with} \textit{confusion}. (Manner \& Mood) \\ 
 His band \textit{combines} professionalism \textbf{with} humor. (Accompanier) \\ 
 He \textit{washed} a small red teacup \textbf{with} \textit{water}. (Means)\\ \hline
\end{tabular}}
\caption{Examples showing polysemous behavior of \emph{with}}
\label{tab:with_senses}
\end{table}

A  PSD algorithm that {\em efficiently scales to a large  corpus}  naturally paves the way for distributed representations  of the preposition senses:   we enrich the corpus  with sense-specific information of prepositions using our PSD algorithm. Next, we repurpose an off-the-shelf word representation algorithm (Word2vec \cite{mikolov2013distributed}) to relearn word representations with the key aspect that the length of the context  surrounding prepositions is crucially reduced. Sense-specific  preposition representations thus learnt are strongly validated using  intrinsic evaluation tasks  on datasets derived from standard benchmarks and by comparing them with their monosemous (i.e., original Word2vec) representations.

Our experiments reveal two curious properties exhibited by sense-specific preposition representations; they encode {\em semantic relations}  and {\em aid paraphrasing of phrasal verbs} when used in a simplistic compositional manner. This compositionality brings forth not only the non-trivial amount of semantic information encoded in prepositions, but also suggests that it can be harnessed using basic algebraic  operations on word representations.
We summarize our contributions below:\\
\textbf{Resource-independent Disambiguation}: We  rely only on a set of trained word representations and not any external linguistic resource --  almost all prior approaches have included at least POS tagging and dependency parsing. We are comparable to, or better than, state-of-the-art on two standard benchmarks. \\
\textbf{Preposition Sense Representation Learning}: To the best of our knowledge, this is the first work on preposition sense representation. The power of our sense representation is reflected by the finding  that the   embedding of `in' in the sense of ``things enclosed", is captured via the linear-algebraic relationship:  in + America $\sim$ American, while the  global representation of `in' fails in such tasks. Again, using a sense-specific word embedding of `for' yields a paraphrase of the phrasal verb, `fight for' to be `defend,' derived via a simple additive model of composition.

A key contribution of this work is in the selectional aspects of the context that best represent the sense of a preposition, where we match  classical ideas from linguistics with the appropriate geometry of word embeddings;  this is discussed next. 


\begin{table*}[ht!]
\centering
\begin{tabular}{|c|c|c|c|c|c|c|}
\hline
\multirow{2}{*}{System} & \multirow{2}{*}{\begin{tabular}[c]{@{}c@{}}State-of-art \\ \cite{hovy2011models}\end{tabular}} & \multicolumn{5}{c|}{$k$-means clustering}  \\ \cline{3-7}  & & \begin{tabular}[c]{@{}c@{}} average \end{tabular} & \begin{tabular}[c]{@{}c@{}} ($\ell,r$) \end{tabular} & \begin{tabular}[c]{@{}c@{}} ($\ell,i$) \end{tabular} & \begin{tabular}[c]{@{}c@{}} ($r,i$) \end{tabular} & \begin{tabular}[c]{@{}c@{}} ($\ell,r,i$) \end{tabular} \\ \hline
Accuracy & $0.56$ & 0.555 & 0.561 & 0.565 & 0.534 & $\mathbf{0.584}$ \\ \hline
\end{tabular}
\caption{Performance of the unsupervised PSD compared with the state-of-the-art. $(\ell$,inter), ($\ell,r)$ and $(r$,inter) correspond to feature ablation results.}
\label{tab:unsupervised_psd}
\end{table*}

\begin{table*}[htbp]
\centering
\begin{tabular}{|c|c|c|c|c|c|c|c|c|c|c|}
\hline
 & \multicolumn{5}{c|}{SemEval Dataset} & \multicolumn{5}{c|}{OEC dataset} \\ \hline
Feature Type & \begin{tabular}[c]{@{}c@{}} average \end{tabular} & ($\ell,r$) & ($\ell,i$) & ($r,i$) & ($\ell,r,i$) & \begin{tabular}[c]{@{}c@{}} average \end{tabular} & ($\ell,r$) & ($\ell,i$) & ($r,i$) & ($\ell,r,i$) \\ \hline
SVM & 0.712 & 0.765 & 0.775 & 0.700 & 0.782 & 0.305 & 0.330 & 0.333 & 0.325 & 0.351 \\ \hline
MLP & 0.712 & 0.758 & 0.780 & 0.704 & 0.777 & 0.322 & 0.353 &  0.353 & 0.347 & 0.375 \\ \hline
Weighted $k$-NN & 0.731 & 0.781 & 0.792 & 0.733 & \textbf{0.804} & 0.329 & 0.341 &  0.380 & 0.367 &  \textbf{0.400} \\ \hline
\end{tabular}
\caption{Supervised disambiguation on SemEval and OEC Datasets.}
\label{tab:supervised_psd}
\end{table*}

\section{Preposition Sense Disambiguation}
\label{sec:psd}
The key intuition behind our sense disambiguation approach is the modern descriptive linguistic view \cite{huddleston1984introduction,decarrico2000structure}: prepositional sense in any sentence is  driven by both its {\em attachment} and its {\em complement}; classical prescriptive linguistics had focused only on the latter \cite{beal}, pp.\ 110,  \cite{cobbett}, pp. 16,  \cite{lowth}, pp.\ 8, 91.   

An example is in Table~\ref{tab:with_senses}: italic words  determine the sense of ``with''. In the first sentence, `confusion' to the right of the preposition (i.e., ``right context") is the complement of `with', from which we infer that `with' encodes the sense of  `manner'. In the second sentence, the accompanier sense of `with'  is because of its governor, the verb `combine' (i.e., the left context). In the last sentence, the sense of `with' is `by means of' and is determined by {\em both} the verb in its left context and the argument in its right  context. Consider a new sentence with changed right context:  `He washed a small cup with a handle.' Here `with'  functions as an attribute. Again, changing its left context we get the sentence `He asked for a small cup with water', where `with' serves as an attribute instead of encoding the sense of means.

That the {\em left} and {\em right} contexts and their {\em interplay} are critical to prepositional sense disambiguation is also well established in the literature \cite{hovy2011models,litkowski2007semeval}. Our key intellectual contribution is in matching these linguistic properties to appropriate {\em geometric} objects within the vector space of word embeddings; the word embeddings are borrowed off-the-shelf -- this work uses word2vec exclusively. We describe this next, focusing first on the left context, next on the right context and then on their interplay. 



\noindent \textbf{Left context feature} $v_{\ell}$ is the average of the vectors of the left $k_{\ell}$ words (here $k_{\ell}$ is a parameter roughly taking values 1 through 4). This simple geometric operation is motivated by  recent works \cite{faruqui2015retrofitting, kenter2016siamese, yu2014deep} representing a sentence by the average of its constituent words robustly and successfully in a variety of downstream settings. Although prior work \cite{hovy2010s} points out that fixed window sizes are insufficient, when compared to using specific syntactic features (example: POS tagging and dependency parsing--common techniques in prior works), we will see that the semantic information embedded in word vectors largely compensates for  this limitation. 

\noindent\textbf{Right context feature} $v_{r}$ is the average of the vectors of the right $k_{r}$ words (here $k_{r}$ is a parameter roughly taking values 1 through 4). This is identical to the method adopted for the left context. 

 \noindent\textbf{Context-interplay  feature} $v_{\text{\rm inter}}$ is the vector {\em closest}  to both the subspace spanned by the left context word vectors and the subspace spanned by the right context word vectors. This geometric representation appears crucial to capture the prepositional-sense when the interplay between the contexts matters decisively, as seen empirically in our extensive experiments. This feature represents one of the key findings of this paper. 
Let $v_{i}^{\ell}$ and $v_{j}^{r}$ be left and right context word vectors respectively. A precise mathematical definition of $v_{\text{\rm inter}}$  is below: 
 \vspace{-0.1in}
\begin{align*}
v_{\text{inter}} = &\arg\min\limits_{v:\lVert v\rVert_{2}=1}\left(\right.\min_{a_1\ldots a_{k_{\ell}}}\lVert v-\sum\limits_{i=1}^{k_{\ell}}a_{i}v_{i}^{\ell}\rVert_{2}^{2} \\
&+\min\limits_{b_1,\ldots b_{k_r}}\lVert v-\sum\limits_{j=1}^{k_{r}}b_{j}v_{j}^{r}\rVert_{2}^{2}\left)\right.
\end{align*}
\vspace{-0.05in}
These three feature vectors, $v_{\ell}$, $v_{r}$ and $v_{\text{inter}}$, are  used in both unsupervised and supervised preposition sense disambiguation.

\textbf{Unsupervised learning} of  senses of a given preposition is conducted by {\em clustering} the 3 feature vectors harnessing the very vast number of instances of each preposition in the large Wikicorpus (here we fix $k_\ell = k_r = 2$ and use standard $k$-means clustering). If the features do capture the prepositional sense efficiently, then a majority of the same-sense instances belong to the same cluster, which is now  represented by the dominant label of its instances. 

\noindent\textbf{Supervised learning} of the senses  using the three feature vectors is  readily conducted based on the  training examples provided in  benchmark  PSD   datasets.   We do this using the  
standard  support vector machines (SVM) \cite{cortes1995support}, multilayer perceptron (MLP) \cite{glorot2010understanding} and weighted $k$-nearest neighbor ($k$-NN) \cite{andoni2006near} classifiers. Each of these allows potentially different  weighting of the three features in a context  dependent way. The parameters are tuned  to maximize the disambiguation accuracy on the development set provided in the benchmark PSD datasets.  These experiments are discussed in detail next.

\section{Experiments on Sense Disambiguation}
\label{exp:psd}

\begin{table*}[]
\centering
\begin{tabular}{|c|c|c|c|}
\hline
Dataset & System & Resources & Accuracy \\ \hline
\multirow{5}{*}{SemEval} & Our system & English corpus & 0.81 \\ \cline{2-4} 
 & \begin{tabular}[c]{@{}c@{}} \cite{litkowski2013preposition}\end{tabular} & \begin{tabular}[c]{@{}c@{}}lemmatizer, dependency parser, WordNet\end{tabular} & {\bf 0.86} \\ \cline{2-4} 
 & \begin{tabular}[c]{@{}c@{}} \cite{srikumar2013modeling}\end{tabular} & dependency parser, WordNet & 0.85 \\ \cline{2-4} 
 & \begin{tabular}[c]{@{}c@{}} \cite{gonen2016semi}\end{tabular} & \begin{tabular}[c]{@{}c@{}}multilingual corpus, aligner,\\ dependency parser\end{tabular} & 0.81 \\ \cline{2-4} 
 & \begin{tabular}[c]{@{}c@{}} \cite{ye2007melb}\end{tabular} & \begin{tabular}[c]{@{}c@{}}chunker, dependency parser, \\ named entity extractor, WordNet\end{tabular} & 0.69 \\ \hline
\multirow{2}{*}{OEC} & Our system & English corpus & {\bf 0.40} \\ \cline{2-4} 
 & \begin{tabular}[c]{@{}c@{}} \cite{litkowski2013preposition}\end{tabular} & \begin{tabular}[c]{@{}c@{}}lemmatizer, dependency parser,  WordNet\end{tabular} & 0.32 \\ \hline
\end{tabular}
\caption{Preposition Disambiguation Performance Comparison on SemEval and OEC dataset}
\label{tab:supervised_psd_comparison}
\end{table*}


The PSD algorithms were validated using two publicly available datasets derived from TPP. \\
The \textbf{SemEval Dataset} consisting of $34$ prepositions instantiated by $24,663$ sentences covering 332 senses. Among them, $16,557$ sentences are used as training instances (\textbf{semtrain}) and $8096$ sentences are test instances (\textbf{semtest}) for the preposition disambiguation task.\\
The \textbf{OEC dataset} consists of $7,650$ sentences collected from the Oxford English Corpus. Since these sentences included more prepositions than those in the SemEval dataset, we chose $3,587$ sentences that included the same 33 prepositions as used in the SemEval task.

\noindent\textbf{Word embeddings}. The word embeddings we used in our experiments were trained on the English WikiCorpus with Word2Vec CBOW model \cite{mikolov2013distributed},  with dimension  300. 

\noindent\textbf{Unsupervised} PSD is conducted by clustering the SemEval dataset's training instances using $k$-means.   In the evaluation phase, {each test instance was assigned to the closest cluster, and its sense was the dominant training sense within this cluster. For a fair comparison with the state-of-the-art unsupervised technique, we report the disambiguation accuracy on \textbf{semtest}} as shown in Table~\ref{tab:unsupervised_psd}, a new state-of-the-art result. 

\noindent{\bf Supervised} PSD is conducted by first  conducting a 80/20 split of \textbf{semtrain} into  training and  development sets. Disambiguation accuracy  calculated on both \textbf{semtest} and OEC datasets are reported in  Table~\ref{tab:supervised_psd}, using  standard off-the-shelf classifiers. 
We used the SVM classifier with a linear kernel and  its penalty parameter $C$ as a tunable parameter, 
 the MLP classifier with one hidden layer, and  the number of neurons as a tunable parameter, and the $k$-NN classifier (weighted $k$-NN), with the number of nearest neighbors  and the feature weights as tunable parameters; all tunable parameters were tuned using the development set.  Additionally, the context window sizes $k_{\ell}$ and $k_r$ were parameters for all the three classifiers, each tuned on the development set.

\noindent\textbf{Baseline}. 
Recent works have shown that the  average word embedding serves as a good representation  of the compositional sentential semantics  \cite{faruqui2015retrofitting, kenter2016siamese, yu2014deep}, and this single feature -- 
the  average of all context word vectors (both to the left and the right) -- serves as a natural baseline.  

\noindent\textbf{Results}. In both the unsupervised and supervised disambiguation settings, the best performance is achieved by using {\em all three} features, $v_{\ell}$, $v_{r}$ and $v_{i}$. 
As summarized  in Table~\ref{tab:unsupervised_psd}, our unsupervised method achieves a  $2.4\%$ 
improvement (4.2\% relative) over state-of-art \cite{hovy2011models}. 

The results  in the supervised setting, tabulated in  Table~\ref{tab:supervised_psd} reveal that the weighted  $k$-NN classifier performs best. Denoting    left, right and interplay features by $\ell,r,i$ respectively,  Table~\ref{tab:unsupervised_psd} and ~\ref{tab:supervised_psd} report our  experimental results using only  subset combinations of these features on the two disambiguation tasks. 

An ablation analysis of the features reveals that  the context-interplay  feature  is most beneficial to the model when  testing  on the OEC dataset, but on the SemEval dataset, the left context feature appears to be the most beneficial.  A likely explanation to this behavior is that {several instances in \textbf{semtrain} and \textbf{semtest} share the governors the prepositions attach to. Hence the left feature with the governor information helps disambiguation on \textbf{semtest}. The governors and complements in OEC instances differ from those in \textbf{semtrain}. Therefore, the context-interplay  feature provides  more general context information than provided by the left and right context features by themselves for sense disambiguation on the OEC dataset.}

A side-by-side comparison of the performance of our supervised approach  with related prior approaches is shown in Table~\ref{tab:supervised_psd_comparison}. From the table we note that the accuracy of our system  was significantly better than that of the best PSD system in SemEval 2007 ($11\%$ higher accuracy), and $7.5\%$ (absolute) higher on the OEC dataset. It is noteworthy that while  \cite{litkowski2013preposition} fared better than our system with the SemEval data, our system outperformed \cite{litkowski2013preposition}  on the OEC dataset. It is also noteworthy that we achieve performance comparable to the recent work \citep{gonen2016semi} which also used word embedings, but had access to  a multilingual translation corpus (and  linguistic tools).  Again, we note that our  performance is achieved with  complete non-reliance on linguistic resources.



\section{Preposition Sense Representation}
\label{sec:representation}
Standard embedding methods do not account for the inherent polysemy in words -- this is exacerbated in the context of  prepositions.  Indeed, to the best of our knowledge, no linguistic properties of the standard embeddings (say, word2vec \cite{mikolov2013distributed} or GloVe \cite{pennington2014glove}) are known for preposition vectors. Recent works that learn sense-specific embeddings inherently use the distinct ``topics" the senses of a given word can take (example: \cite{rothe2015autoextend} explicitly uses Wordnet senses) and have only been validated with respect to nouns and verbs.   

In this work, we provide  the first sense-specific prepositional representations and validate them by creatively repurposing datasets meant for other tasks. This is the focus of this section. 
Toward this, we used the trained $k$-NN  classifier (described in Section~\ref{sec:psd})  to disambiguate {\em each}  preposition token in the large WikiCorpus. Now each preposition instance in the corpus has a sense-label.  We then used Word2Vec \cite{mikolov2013distributed}  to re-learn word embeddings  on the preposition-sense-tagged corpus; this time we arrive at sense-specific embeddings of prepositions. 

The sense-specific representations  are readily interpreted in terms of the extensive-resources of TPP -- a detailed description of our sense representations and their connections to  TPP senses can be found in Tables~\ref{tab:in_senses},\ref{tab:over_senses},\ref{tab:of_senses},\ref{tab:for_senses},\ref{tab:with_senses_appendix} of the supplementary material, including the words nearest to the preposition sense  and corresponding  example sentences for five common prepostions: {\em in, over, for, or, with}. 

Below, we 
validate the quality of the sense representations  in two tasks, where prepositional senses play an important semantic role: (a)  semantic analogy task and (b)  paraphrasing task.

\begin{table}[]
\centering
\resizebox{0.5\textwidth}{!}{
\begin{tabular}{|c|c|c|c|}
\hline
Embedding & Global & Sense & Difference \\ \hline
Capital-country & 0.17 & 0.54 & 0.95 \\ \hline
City-state & 0.32 & 0.67 & 0.91 \\ \hline
Nationality-adjective & 0.73 & 0.85 & 0.95 \\ \hline
\end{tabular}}
\caption{Accuracy on relation approximation}
\label{tab:prep_relation}
\end{table}

\subsection{Preposition senses as relations}
 \textbf{Task}. Prepositions indicate a relation between the noun or pronoun and another word, which may be a verb, an adjective, or another noun or pronoun \cite{huddleston2002cambridge}.  
 While previous studies have found that simple arithmetic operations between word vectors capture the relation between word pairs fairly well \cite{mikolov2013distributed,mikolov2013linguistic}, in this study we explored the ability of sense-specific preposition embeddings to encode  two {\em noun-noun } relations and one {\em noun-adjective}  relation. 
 The rationale here is explained with an example. 
 
 Prepositions such as  `in' and `from' encode a spatial relation (\emph{in America}) and hence the location sense of these prepositions could potentially capture the  nationality relation that ${\text{\emph{in America}}} \approx {\text{\emph{{American}}}}$.  If the prepositional sense embedding can indeed capture this spatial relation, then the adjective can  be predicted from the country via the addition operation as follows. Let a particular sense embedding of `in' be $v_{\text{in}}^{\text{sense}}$. Given the country, we predict the nationality-adjective by finding the nearest word of $v_{\text{adjective}}^{\text{sense}} \approx v_{\text{in}}^{\text{sense}}+v_{\text{country}}$.   Likewise,  we explored whether country names (resp.\   state) could also be predicted from their capital (resp.\ city) names via the addition operation.  
 
 \noindent\textbf{Datasets:} We use the popular semantic analogy datasets \cite{pennington2014glove} and focus on the following three relations:  
(1) \textbf{capital-world} with  $116$  (capital, country) pairs, e.g., (Cairo, Egypt);  
(2) \textbf{city-in-state} with $69$ (city, state) pairs, e.g., (Houston, Texas);  
(3) \textbf{gram6-nationality-adjective} with $41$  (country, nationality-adjective) pairs, e.g., (Albania, Albanian).

\noindent\textbf{Baselines}. (a) Lower baseline:   the ``global" preposition embedding (i.e., original word2vec representation),  $v_{\text{in}}^{\text{global}}$, is one baseline in our experiments, e.g., we predict the adjective through $v_{\text{adjective}}^{\text{global}}\approx v_{\text{in}}^{\text{global}}+v_{\text{country}}$.

(b) Upper baseline: since the difference between two words in the first analogy pair  are  shown to efficiently capture  the relation, we took the average difference vector among all word pairs in the dataset corresponding to a relation,  as the second baseline, $v^{\text{diff}}$. The adjective is then predicted via $v_{\text{adjective}}^{\text{diff}}\approx v^{\text{diff}}+v_{\text{country}}$.

We used \emph{from} to approximate capital-country relation, and  \emph{in} for city-state and nationality-adjective relations.  For each relation, we evaluated the outcome of adding the preposition vector to the base word by checking if the answer occurs among the closest three words of the sum vector. For example, if \emph{American} appears among the closest words to the sum of the vectors of \emph{America} and \emph{in}, we considered it to be a correct approximation.

\textbf{Results}.  Table~\ref{tab:prep_relation} reports the accuracy of finding the target word (country, state or adjective) in the top 3 neighbors corresponding to the use of the global embedding, the sense-specific embedding of (`in' and `from') and the difference embedding. 
The accuracy achieved by using the preposition sense representation is significantly close to that of the difference embedding compared to the global representation. This shows that the sense representation is good at  approximating the relations between capital-state, capital-country and country-nationality. 

\begin{table*}[htbp!]
\centering
\begin{tabular}{|c|c|c|c|c|}
\hline
\multirow{2}{*}{sentence} & \multirow{2}{*}{phrasal verb} & \multicolumn{3}{c|}{paraphrasing} \\ \cline{3-5} 
 &  & \begin{tabular}[c]{@{}c@{}}sense \end{tabular} & \begin{tabular}[c]{@{}c@{}}global \end{tabular} & \begin{tabular}[c]{@{}c@{}}simplex \end{tabular} \\ \hline
\begin{tabular}[c]{@{}c@{}}The teaching is \textbf{carried on} in the form of folklore.\end{tabular} & carried on & \textbf{conducted} & laid & placed \\ \hline
\begin{tabular}[c]{@{}c@{}}he \textbf{brought in} new ideas  in the discussion.\end{tabular} & brought in & \textbf{introduced} & came & came \\ \hline
\end{tabular}
\caption{Paraphrasing of Phrasal Verbs}
\label{tab:vp_paraphrasing}
\end{table*}

\subsection{Preposition senses aid paraphrasing}

\textbf{Task}. {Prepositions encode  non-trivial semantic information. For example, \emph{switch on} and \emph{switch off} show opposite meanings owing to the prepositions that follow the  common verb \emph{switch}. 
Another setting in which we validate the sense-specific  preposition representation is by understanding its role in  phrasal verbs. 

Specifically, our goal is to infer the meaning of verb-particle constructions (VPC)--a head verb with one or more obligatory particles -- in the form of intransitive prepositions (e.g.,  \emph{hand in}). We focused  exclusively on prepositions serving as particles due to their high productivity, and mainly consider compositional VPCs \cite{mccarthy2003detecting,bannard2003statistical}. 
This allows us to  highlight the value of the vector representation of the preposition sense in terms of playing a non-trivial role in phrasal verb semantics  \cite{brinton1985verb}.%
 
 }
 


\noindent\textbf{Experiments}. We explore the task of inferring the meaning of the phrasal verb  from its components, i.e., the verb and preposition sense representation,  casting this as a lexical paraphrasing task of finding one word that captures the meaning of the VPC (e.g., {\em climb down = descend}).

\noindent\textbf{Dataset}. Because a dataset for paraphrasing of VPCs was not available, we created a dataset \footnote{availabl at:\url{ https://github.com/HongyuGong/PrepositionDisambiguationAndRepresentation.git}} . It  consists of $91$ phrasal verbs,  extracted from the VPC datasets in \cite{baldwin2005deep},   \cite{mccarthy2003detecting} and the online Oxford dictionary\footnote{\url{https://en.oxforddictionaries.com}}.  
Given that the meaning of VPCs is context sensitive (as discussed in \cite{gong2016geometry} for example), we provide three sentences for each VPC to ascertain the paraphrase, while ensuring that the VPC has the same sense in all three sentences. 

For each VPC instance, we first  disambiguated the preposition sense in the given context using the supervised method described in Section~\ref{sec:psd}. Because  the meaning of a compositional phrase  can be inferred from the meaning of its component words,   we approximate the word representation of a VPC as the sum of the vectors of its verb and its preposition. Thus, we have,  $v_{\text{vp}} = v_{\text{verb}}+v_{\text{prep}}$. We consider such an approximation under three settings:  \\
(1) {\em Global embedding baseline}: In this simplistic compositional model of the phrasal verb, we add the verb and the global preposition embedding to approximate the phrasal verb embedding, i.e., $v_{\text{vp}}^{\text{global}} = v_{\text{verb}}+v_{\text{prep}}^{\text{global}}$;  \\ 
(2) {\em Simplex embedding baseline}: Here the assumption is that the verb alone is contributing to the meaning of the phrasal verb. Hence, we use the verb embedding alone, to approximate the phrasal verb embedding, i.e., $v_{\text{vp}}^{\text{simplex}} = v_{\text{verb}}$;  \\ 
(3) {\em Sense-specific embedding}: Here we use our sense-specific preposition embedding to yield $v_{\text{vp}}^{\text{sense}} = v_{\text{verb}}+v_{\text{prep}}^{\text{sense}}$

For each approximate phrasal embedding ($v_{\text{vp}}^{\text{sense}}$, $v_{\text{vp}}^{\text{global}}$,$v_{\text{vp}}^{\text{simplex}}$), we list the 
nearest three verbs (excluding the verb in the phrase) as its paraphrase. Here, the distance is measured in terms of the cosine similarity between the word vectors. Examples of phrasal verb paraphrasing are shown in Table~\ref{tab:vp_paraphrasing}. In the sentence, ``The teaching was \textbf{carried on}  in the form of folklore'', the nearest neighbor of \emph{carry on} is \emph{conduct} using the preposition sense embedding, \textit{laid} using the global embedding and \textit{placed} with the simplex verb. 

Two human annotations set the gold standard  for  whether the  
 paraphrase is valid or not (for  polysemous verbs, we consider the verb as a valid paraphrase if it conveys the meaning in any of its senses). The agreed upon annotations   constitute the dataset.   
We use accuracy as evaluation metric, which is the percent of phrasal verbs with a valid paraphrase among three candidates. A more detailed evaluation is in the supplementary material.

\begin{table}[]
\centering
\resizebox{0.46\textwidth}{!}{
\begin{tabular}{|c|c|c|c|}
\hline
Embedding & Global & Simplex & Sense \\ \hline
Accuracy & 0.44 & 0.44 & \textbf{0.73} \\ \hline
\end{tabular}}
\caption{Accuracy on Phrasal Verbs Paraphrasing.}
\label{tab:vp_para_at_3}
\end{table}

 

\noindent\textbf{Results}. We report the results in Table~\ref{tab:vp_para_at_3}, where 
we notice that paraphrasing with the preposition sense embedding has a much higher precision  than the two baselines. This
 validates the sense-specific preposition embedding and suggests that its use helps  automatic paraphrasing  of  VPCs.

 Comparing the different phrasal verb approximation  methods  on an instance-by-instance basis yields a closer view of the results. Of the 91 phrasal verbs, there were 31 instances where a sense-based approximation was better than  that using a global-embedding, 32 instances where sense-based was better than simplex, and 19 instances where sense-based was better than both global and simplex. This shows that the role of sense-specific preposition embeddings in  capturing the meanings of  phrasal verbs is non-trivial.


\noindent{\bf  Sense embeddings outperform simplex} ones  in instances where:  (a) prepositions are important in aspectual phrases (where the particle provides the verb with an endpoint, suggesting that the action described by the verb is performed completely, thoroughly or continuously), e.g., "go against"; (b) prepositions help disambiguate the verb, e.g., "carried" has multiple senses, sense 1: support the weight of something, sense 2: Assume or accept (responsibility or blame). In vector representation, "carried" is close to "laid", "wiped" and "phased",  while sense "on" drives "carry on" much closer to "conducted".

\noindent{\bf   Sense embeddings outperform global} ones since the latter   only represent the semantics of dominant sense while sense embedding is better at capturing the true sense. For example, the global embedding of \textit{down} is close to \textit{destroyed} and \textit{crashed}, and thus in phrase \textit{put down}, global method gives paraphrases such as  \textit{slammed} and \textit{snapped}. Sense embedding provides its sense of ``downward direction'', and gives the paraphrases  \textit{laid} and \textit{tossed}.

\noindent {\bf Sense embeddings encode phrasal verb} semantics even though the preposition in the phrasal verb has lost its functional aspect;  we see that computationally (and in a vector space), the sense-tagged preposition remains inside a phrasal verb.  This is more pronounced in compositional phrasal verbs and  in aspectual  ones, and  less so in idiomatic ones    (see 
Section~\ref{app:paraphrase} for a discussion \cite{villavicencio2006verb}).

\section{Discussion}
{\bf Resource-independence}: Previous approaches to PSD relied on a dependency parser to extract words modified by a preposition and those that the preposition modifies. 
In general, these words  occur in the preposition's local context. We have allowed the context window to be a tunable parameter so that the classifier can learn to cover informative words in the context, and thus effectively captures the dependency  information in a resource-independent fashion. 


\noindent {\bf Novel context feature}: The context averaging approach, which disregards context word order, suffers in accuracy compared to models that use left and right context words.  This indicates that information about the order relative to the preposition is useful in preposition disambiguation, since the left (resp.\ right) context generally corresponds to attachment (resp.\ complement) information. Additionally, 
 our use of the context-interplay  feature  combines the information on {\em both} sides of the preposition to infer its  underlying sense. Suppose \emph{a cup of medicine}, \emph{professor of humanity} and \emph{professor of mathematics} are in the training corpus, and senses of preposition \emph{of} are `contents', `possessor' and `field'.
Given a test instance \emph{professor of medicine}, it would be hard for the method with only the left or the right feature to decide the preposition sense since the test instance has the same word as each of the training instance, and their features in these two baselines are similar. 
However, the interplay vector in \emph{professor of medicine} is closer to that in ``professor of mathematics"  than to other two training instances. The interplay feature prompts that \emph{of} in test instance refers to a field (or species) instead of contents or possessor.  

\noindent  {\bf Data-driven insights into context dependence}: Knowing the weights on the context features in our supervised PSD model, the weighted $k$-NN,  we can infer the extent to which  prepositions rely on the complement and the attachment.  
For example, we found that  in the case of the prepositions \emph{behind} (occurring in, ``shut behind her'', ``dip behind clouds''), \emph{to} (e.g., ``testify to the depth'', ``mumbling to himself''), and \emph{with} (e.g., ``amalgamated with her old school'',  and ``rub with bare hands''), the verbs they attach to strongly influence their sense. 
For other prepositions such as \emph{during} (e.g., ``during the incident'', ``during his lifetime'', ``during the day'') and \emph{on} (e.g.,  ``on his hands'', ``on the ground''),  the complement  has more influence on the senses.  

\noindent {\bf Sense encodes relations}: 
Sense-specific representations  outperform the global preposition representation in terms of encoding {\em semantic relations}--thus prepositional  sense-specificity captures the encoded semantics better than its sense-generic version. Working with the small VPC dataset and the simplistic model of compositionality, we interpret the results as positive indicators of the viability of using sense-specific prepositional embeddings to paraphrase VPCs. We observe that in the case of {\em light verbs}, whose meaning is determined by the particles they combine with, (e.g., \textit{come down} $\sim$ \textit{fall}), a valid paraphrase is found in the top 3 candidates when the sense-specific representation is used, and not when the simplex or the global representation is used.

As pointed out in \cite{navigli2006meaningful}, a potential limiting factor of the sense-specific representation could be the fine-grained sense distinctions in the training set. Future work could explore preposition sense representation learned from  a coarse-grained traininng set.

\section{Related Works}
\label{sec:relatedWork}



\noindent \textbf{Preposition Sense Disambiguation}:
Preposition disambiguation has been explored on the SemEval dataset via various methods and external resources (part of speech taggers, chunkers, dependency parsers, named entity extractors, WordNet based supersense taggers and semantic role labelers) since 2007 \cite{yuret2007ku,ye2007melb,tratz2009disambiguation,hovy2011models,popescu2007irst,tratz2011fast,srikumar2013modeling}. 

More recently, \citet{gonen2016semi} use word embeddings and other resources including a multilingual parallel corpus  processed using sequence to sequence neural networks for preposition disambiguation and achieve  an accuracy within 5\% of the state-of-the-art, which includes  \cite{litkowski2013preposition,hovy2010s,srikumar2013modeling}. We note that we achieve the comparable performance as  \cite{gonen2016semi} using  {\em only} word embeddings.  

\noindent\textbf{Preposition Representation}:  Word embeddings such as  Word2vec \cite{mikolov2013distributed} and GloVe \cite{pennington2014glove} have been widely recognized for their ability to  capture linguistic regularities (including syntactic and semantic relations). 
On the other hand, no linguistic property of their prepositional embeddings are known; to the best of our knowledge, we propose the first sense-specific prepositional embeddings and demonstrate their linguistic regularities. Distantly  related is  \cite{hashimoto2015learning},  which learns embeddings of prepositions  acting as verb adjuncts   by tensor factorization of a predicate matrix. 
Similarly, \citet{belinkov2014exploring} explore  the use of preposition representations optimized for  the task of prepositional phrase attachment, but do not  analyze  the semantic contribution  or sense-specificity of preposition embeddings.


\noindent{\bf Sense-specific Embedding}: Recent works have  proposed polysemy disambiguation,  using external resources such as  Wordnet \cite{rothe2015autoextend} or in an unsupervised way  \cite{arora2016linear,neelakantan2015efficient}; both the unsupervised approaches are limited in the number of senses they can represent (about 4) and are  validated for  only nouns and verbs.  The approach of  \cite{neelakantan2015efficient} is roughly similar to our baseline 
 method using the average context vector. 

\section{Conclusion}
This paper encodes attachment and complement properties of prepositions into context features, disambiguating senses of preposition. The method  relies on no external resources (all prior works use at least a dependency parser), and performs very well on two standard PSD datasets. The disambiguation readily scales to a  large corpus and the resulting sense-specific representations have been shown to  capture lexical relationships and aid phrasal paraphrasing.  Evaluating the utility of  the  preposition representations in downstream  NLP applications (specifically question-answering) is left to future  work.

\newpage
\bibliography{acl2017}
\bibliographystyle{acl_natbib}

\newpage

\appendix

\onecolumn
\begin{center}
	\Large Supplementary Material: Prepositions in Context
\end{center}
\section{Preposition Sense Representation}
\subsection{Word similarity task}
We learn preposition sense representations, and explore the semantic information they carry. A  straightforward approach is to examine the nearest neighboring words, corresponding to each sense of each representation. Our results on this exploration for five exemplar prepositions are here: \emph{in} (Table~\ref{tab:in_senses}), \emph{over} (Table~\ref{tab:over_senses}), \emph{for} (Table~\ref{tab:for_senses}), \emph{of} (Table~\ref{tab:for_senses}), and \emph{with} (Table~\ref{tab:with_senses_appendix}). We enumerate senses for a preposition in each table, and also provide their closest words, TPP semantic type and example sentences for each sense.
 
\begin{table*}[htbp!]
\centering
\resizebox{1.0\textwidth}{!}{
\begin{tabular}{|c|c|c|c|c|c|c|}
\hline
\begin{tabular}[c]{@{}c@{}}sense\\ number\end{tabular} & 1 & 2 & 3 & 4 \\ \hline
\begin{tabular}[c]{@{}c@{}}closest\\ words\end{tabular} & \begin{tabular}[c]{@{}c@{}}backwards, reverse, angles, \\ diagonal, between, forward\end{tabular} & \begin{tabular}[c]{@{}c@{}}wearing, dress, hats, dresses, \\ trousers, sleeves, pants, jacket\end{tabular} & \begin{tabular}[c]{@{}c@{}}back, inside, underneath, \\ from, into, where, onto\end{tabular} & \begin{tabular}[c]{@{}c@{}}where, near, from, at, \\ southern, northern,during\end{tabular} \\ \hline
example & \begin{tabular}[c]{@{}c@{}}in all directions,\\ move in, differ in\end{tabular} & dress in black, in leather, in size & \begin{tabular}[c]{@{}c@{}}in the mail, \\ in most cases, \\ in confined space\end{tabular} & \begin{tabular}[c]{@{}c@{}}in military aircraft, \\in the UK, in Argentina\end{tabular} \\ \hline
\begin{tabular}[c]{@{}c@{}}TPP\\ sense\end{tabular} & Manner\_or\_Degree & VariableQuality & ThingEntered & ThingEnclosed \\ \hline
\begin{tabular}[c]{@{}c@{}}sense\\ number\end{tabular} & 5 & 6 & 7 & 8 \\
\hline
\begin{tabular}[c]{@{}c@{}}closest \\ words\end{tabular} & \begin{tabular}[c]{@{}c@{}}until, during, subsequently, \\following, after, late, since\end{tabular} & \begin{tabular}[c]{@{}c@{}}university, graduate, college, \\teaching, faculty, school\end{tabular} & \begin{tabular}[c]{@{}c@{}}economic, systematic, \\growth, technological\end{tabular} & \begin{tabular}[c]{@{}c@{}}wearing, dressed, costume, \\ wears, clothes, jacket\end{tabular} \\ \hline
example & \begin{tabular}[c]{@{}c@{}}in 1978,  in may 1993,\\ in 2002, in the weeks\end{tabular} & \begin{tabular}[c]{@{}c@{}}in a lecture, \\ in graduate studies,\\ in college\end{tabular} & \begin{tabular}[c]{@{}c@{}}focus in science,\\ growth in sales,\\ vocals in her pieces\end{tabular} & \begin{tabular}[c]{@{}c@{}}in the costume,\\ in the jacket,\\ in a gown\end{tabular} \\ \hline
\begin{tabular}[c]{@{}c@{}}TPP\\ sense\end{tabular} & Timeframe & ProfessionAspect & Attribute & Garment \\ \hline
\begin{tabular}[c]{@{}c@{}}sense\\ number\end{tabular} & 9 & 10 & 11 & 12 \\ \hline
\begin{tabular}[c]{@{}c@{}}closest \\ words\end{tabular} & \begin{tabular}[c]{@{}c@{}}explicitly, interpretation, \\discourse, fundamental, \\notion, principles \end{tabular} & \begin{tabular}[c]{@{}c@{}}prosecutor, prosecution, \\criminal, judicial, justice\end{tabular} & onwards, for, wherein & \begin{tabular}[c]{@{}c@{}}violent, betrayal, bloody, \\ brutal, bitter, fearful\end{tabular} \\ \hline
example & \begin{tabular}[c]{@{}c@{}}in a diagram, in this process,\\ in different ways, in the work\end{tabular} & \begin{tabular}[c]{@{}c@{}}in a constitution, \\in military justice, in court \end{tabular} & \begin{tabular}[c]{@{}c@{}}in computer graphics, \\ in engineering projects,\\ in the war\end{tabular} & \begin{tabular}[c]{@{}c@{}}result in, in custody,\\ involved in, participate in\end{tabular} \\ \hline
\begin{tabular}[c]{@{}c@{}}TPP\\ sense\end{tabular} & Medium & Activity & FramingEntity & Condition \\ \hline
\end{tabular}}
\caption{Senses of Preposition ``in''}
\label{tab:in_senses}
\end{table*}

In Table~\ref{tab:in_senses}, we notice that the nearest neighbors can be the words semantically similar to the given sense. For example, \emph{until}, \emph{during} and \emph{since} are close to TimeFrame sense of \emph{in}. The nearest words might be the governors of a preposition. For example, when \emph{in} carries the sense of Garment, verbs such as \emph{dressed} are close to it. Also, the nearest words can also be complements of this preposition. For example, nouns such as \emph{university}, \emph{college} and \emph{school} are neighbors to \emph{in}'s sense of ProfessionAspect.

\begin{table}[htbp!]
\centering
\resizebox{1.0\textwidth}{!}{
\begin{tabular}{|c|c|c|c|c|}
\hline
\begin{tabular}[c]{@{}c@{}}sense \\ number\end{tabular} & 0 & 1 & 2 & 3 \\ \hline
\begin{tabular}[c]{@{}c@{}}closest\\ words\end{tabular} & \begin{tabular}[c]{@{}c@{}}around, upwards, \\ across, outwards, \\ vertically, surface\end{tabular} & \begin{tabular}[c]{@{}c@{}}crossed, foot, paces,\\ straightened, stretch,\\ spanning, river\end{tabular} & \begin{tabular}[c]{@{}c@{}}about, around, \\ intervening,\\ estimated\end{tabular} & \begin{tabular}[c]{@{}c@{}}onto, grabbed, inside,\\ top, apiece,\end{tabular} \\ \hline
example & \begin{tabular}[c]{@{}c@{}}glace over \\ her shoulder,\\ look over the \\ gymnasium\end{tabular} & \begin{tabular}[c]{@{}c@{}}bridge over \\ the river,\\ leaned over\\ her shoulder\end{tabular} & \begin{tabular}[c]{@{}c@{}}brood over it,\\ pondered over \\ the reply\end{tabular} & \begin{tabular}[c]{@{}c@{}}flung a net over me,\\ drizzling ketchup \\ over his grilled chicken\end{tabular} \\ \hline
TPP sense & ThingsSurveyed & ThingsSurmounted & SubjectConsidered & ThingsCovered \\ \hline
\begin{tabular}[c]{@{}c@{}}sense \\ number\end{tabular} & 4 & \textbf{5} & 6 & 7 \\ \hline
\begin{tabular}[c]{@{}c@{}}closest\\ words\end{tabular} & \begin{tabular}[c]{@{}c@{}}heat, coated, oven,\\ insulation, heaters\end{tabular} & \begin{tabular}[c]{@{}c@{}}spanning, stretch, \\ crosses, spans, \\ longest, reaching\end{tabular} & \begin{tabular}[c]{@{}c@{}}notching, yards,\\ shut-out, scampered\end{tabular} & \begin{tabular}[c]{@{}c@{}}across, traversing,\\ meters, submerged,\\ traversed, inland\end{tabular} \\ \hline
example & \begin{tabular}[c]{@{}c@{}}cook the dumplings \\ over a medium heat,\\ the fighting over England,\end{tabular} & \begin{tabular}[c]{@{}c@{}}crossed one foot \\ over another\end{tabular} & \begin{tabular}[c]{@{}c@{}}discarded the broken \\ spar over the side\end{tabular} & \begin{tabular}[c]{@{}c@{}}clambered over \\ the fallen rocks\end{tabular} \\ \hline
TPP sense & ThingsSurmounted & PlaceSurpassed & ThingDescendedFrom & ThingNegotiated \\ \hline
\begin{tabular}[c]{@{}c@{}}sense \\ number\end{tabular} & 8 & 9 &  &  \\ \hline
\begin{tabular}[c]{@{}c@{}}closest \\ words\end{tabular} & \begin{tabular}[c]{@{}c@{}}scored, unbeaten, \\ against, beating, \\ hat-trick, rule\end{tabular} & \begin{tabular}[c]{@{}c@{}}hit, against, on,\\ broke, smashed\end{tabular} &  &  \\ \hline
example & \begin{tabular}[c]{@{}c@{}}reigned over \\ Bangkok 's economy,\\ rule over it\end{tabular} & \begin{tabular}[c]{@{}c@{}}broke a chair \\ over me\end{tabular} &  &  \\ \hline
TPP sense & ThingControlled & ResistantSurface &  &  \\ \hline
\end{tabular}}
\caption{Senses of Preposition \emph{over}}
\label{tab:over_senses}
\end{table}

For the sense representations of \emph{over} in Table~\ref{tab:over_senses}, we  see that the nearest neighbors are indeed synonymous. For example, \emph{crossed} is close to \emph{over}'s sense of ThingsSurmounted, \emph{about}  close to the sense of SubjectConsidered, \emph{onto} close to the sense of ThingsCovered, \emph{against} close to the sense of ResistantSurface. Senses are interchangeable with synonymous neighbors. ``bridge \emph{over} the river'' is similar to ``bridge \emph{crossed} the river'', ``ponder \emph{over} the reply'' similar to ``ponder \emph{about} the reply'', and ``drizzling ketchup \emph{over} chicken'' similar to ``drizzling ketchup  \emph{onto} chicken'', and ``broke a chair \emph{over} me'' similar to ``broke a chair \emph{against} me''. Besides synonyms, nearest words also include governors and complements specific to \emph{over}'s senses since they co-occur so frequently. Word \emph{broke} is a governor in ``broke a chair over me'', and it is also a nearest neighbor given \emph{over}'s sense as ResistantSurface. In another sentence ``cook the dumplings over a medium heat'',  \emph{heat} is a complement, and also close to \emph{over}, given its sense  as ThingsSurmounted.

\begin{table}[htbp!]
\centering
\resizebox{1.0\textwidth}{!}{
\begin{tabular}{|c|c|c|c|c|c|c|}
\hline
\begin{tabular}[c]{@{}c@{}}sense\\ number\end{tabular} & 0 & 1 & 2 & 3 \\ \hline
\begin{tabular}[c]{@{}c@{}}closest\\ words\end{tabular} & \begin{tabular}[c]{@{}c@{}}within, its, \\constituting,\\ comprising, \\ combined\end{tabular} & \begin{tabular}[c]{@{}c@{}}cups, pot, basket,\\ bucket, bowls\end{tabular} & \begin{tabular}[c]{@{}c@{}}featured, wonderful, \\amazing, \\ shows, laugh\end{tabular} & \begin{tabular}[c]{@{}c@{}}concerning, \\ regarding, furthermore\end{tabular} \\ \hline
example & \begin{tabular}[c]{@{}c@{}}colonies of insects,\\ bunches of grain\end{tabular} & \begin{tabular}[c]{@{}c@{}}a can of milk,\\ the envelop of \\ photographs\end{tabular} & \begin{tabular}[c]{@{}c@{}}a smile of delight,\\ a frown of doubt\end{tabular} & \begin{tabular}[c]{@{}c@{}}purchase of copyright,\\ smuggling of oil\end{tabular} \\ \hline
TPP sense & Whole & Contents & Defining Quality & Recipient \\ \hline
\begin{tabular}[c]{@{}c@{}}sense\\ number\end{tabular} & 4 & 5 & 6 & 7 \\ \hline
\begin{tabular}[c]{@{}c@{}}closest\\ words\end{tabular}  & \begin{tabular}[c]{@{}c@{}}specific, particular,\\ common, generic\end{tabular} & \begin{tabular}[c]{@{}c@{}}measured, maximum, decreases,\\  increases, proportional\end{tabular} & \begin{tabular}[c]{@{}c@{}}triangular, outer, \\ underneath, adjacent\end{tabular} & \begin{tabular}[c]{@{}c@{}}appointed, deputy,\\ treasurer, whose, \\chief, governed\end{tabular} \\ \hline
example & \begin{tabular}[c]{@{}c@{}}a type of,\\ all kinds of\end{tabular} & \begin{tabular}[c]{@{}c@{}}length of 7 cm,\\ \\ an increase of 1\%\end{tabular} & \begin{tabular}[c]{@{}c@{}}top of the slope,\\ face of a man\end{tabular} & \begin{tabular}[c]{@{}c@{}}descendant of humans,\\ sisters of this girl\end{tabular} \\ \hline
TPP sense & Exemplar & ScaleValue & Whole & Possessor \\ \hline
\begin{tabular}[c]{@{}c@{}}sense\\ number\end{tabular} & 8 & 9 & 10 & 11 \\ \hline
\begin{tabular}[c]{@{}c@{}}closest\\ words\end{tabular} &  \begin{tabular}[c]{@{}c@{}}corrupted, enlightened,\\ regards, owes\end{tabular} & \begin{tabular}[c]{@{}c@{}}gender,male, female,\\  adolescent, adult\end{tabular} & \begin{tabular}[c]{@{}c@{}}containing, packaged, \\handmade, contained\end{tabular} & \begin{tabular}[c]{@{}c@{}}acknowledge, asserts,\\ argues, conceived,\\ believes\end{tabular} \\ \hline
example & \begin{tabular}[c]{@{}c@{}}tell of the experience, \\ enlightenment of \\participants\end{tabular} & \begin{tabular}[c]{@{}c@{}}infants of the same age,\\ women of nineteen\end{tabular} & \begin{tabular}[c]{@{}c@{}}stick of furniture, \\ the firms of solicitors\end{tabular} & \begin{tabular}[c]{@{}c@{}}ashamed of,\\ conceived of,\\ know of\end{tabular} \\ \hline
TPP sense  & Object & Age & Constituent & MentalContents \\ \hline
\begin{tabular}[c]{@{}c@{}}sense\\ number\end{tabular} & 12 & 13 & 14 & 15   \\ \hline
\begin{tabular}[c]{@{}c@{}}closest\\ words\end{tabular} & \begin{tabular}[c]{@{}c@{}} died, buried, \\survived, pleaded,\\ dying, arrested\end{tabular} & \begin{tabular}[c]{@{}c@{}}seeming, utter,\\ sincere, religious\end{tabular} & \begin{tabular}[c]{@{}c@{}}consequent, upon, \\regarding, subsequent\end{tabular} & \begin{tabular}[c]{@{}c@{}} emphasize, \\concerning, \\ embodied, \\academic, \\faculty\end{tabular} \\ \hline
example & \begin{tabular}[c]{@{}c@{}}died of a coronary,\\ convict the farmer of \\ pollution\end{tabular} & \begin{tabular}[c]{@{}c@{}}afraid of,\\ religious of,\\ shy of\end{tabular} & \begin{tabular}[c]{@{}c@{}}belief of people, \\presumption of \\the courts \end{tabular}& \begin{tabular}[c]{@{}c@{}}expression of, \\ announcement of \\ a discovery\end{tabular}  \\ \hline
TPP sense & Cause & Concomitant & Possesor & Species   \\ \hline
\begin{tabular}[c]{@{}c@{}}sense\\ number\end{tabular} & 16 & & & \\ \hline
\begin{tabular}[c]{@{}c@{}}closest\\ words\end{tabular} & \begin{tabular}[c]{@{}c@{}}written, novels, adaptations, \\narrated, edited\end{tabular} & & & \\ \hline
example & \begin{tabular}[c]{@{}c@{}}diary of Bob,\\ poetry of Shakespare\end{tabular} & & & \\ \hline
TPP sense & Creator & & & \\ \hline
\end{tabular} }
\caption{Senses of Preposition \emph{of}}
\label{tab:of_senses}
\end{table}

Seventeen senses are enumerated for preposition \emph{of} in Table~\ref{tab:of_senses}. Synonymous words can be found in \emph{of}'s nearest neighbors. For example, \emph{concerning} is close to the sense Recipient, \emph{upon} is close to the sense Possessor, and \emph{containing} is close to the sense Constituent. These neighbors can paraphrase \emph{of}'s corresponding senses . The sentence ``purchase \emph{of} copyright'' can be paraphrased as ``purchase \emph{concerning} copyright'',   ``presumption \emph{of} the courts'' paraphrased as ``presumption \emph{upon} the courts'', and  ``the firms \emph{of} solicitors'' as ``the firms \emph{containing} solicitors''. We again observe that the nearest neighbors also reflect the attachment and complement properties of specific senses. When \emph{of} carries the sense of Contents, words such as \emph{cups}, \emph{pot} and \emph{bowls} are neighbors of the sense Contents. When \emph{of} carries the sense of Cause, governors such \emph{died} and \emph{arrested} are its closest neighbors.

\begin{table}[htbp!]
\centering
\resizebox{1.0\textwidth}{!}{
\begin{tabular}{|c|c|c|c|c|}
\hline
\begin{tabular}[c]{@{}c@{}}sense\\ number\end{tabular} & 0 & 1 & 2 & 3 \\ \hline
\begin{tabular}[c]{@{}c@{}}closest\\ words\end{tabular} & \begin{tabular}[c]{@{}c@{}}purchase, buy, \\ rental, lease\end{tabular} & \begin{tabular}[c]{@{}c@{}}rovers, starting, \\ southend\end{tabular} & \begin{tabular}[c]{@{}c@{}}promotional, showcase,\\ featuring, music, \\ advertise, commercial\end{tabular} & \begin{tabular}[c]{@{}c@{}}generic, terminology,\\ describe, denote,\\ definitions, defined\end{tabular} \\ \hline
example & \begin{tabular}[c]{@{}c@{}}bought it for \$200\\ purchase shares for cash\end{tabular} & \begin{tabular}[c]{@{}c@{}}headed for \\ the bathroom,\\ made for \\ the orchard\end{tabular} & \begin{tabular}[c]{@{}c@{}}feel pity for,\\  be worried for,\\ be embarrassed for\end{tabular} & \begin{tabular}[c]{@{}c@{}}a synonym \\ for coordination,\\ an expression \\ for the energy\end{tabular} \\ \hline
\begin{tabular}[c]{@{}c@{}}TPP\\ sense\end{tabular} & Price & Destination & Beneficiary & Referent \\ \hline
\begin{tabular}[c]{@{}c@{}}sense\\ number\end{tabular} & 4 & 5 & 6 & 7 \\ \hline
\begin{tabular}[c]{@{}c@{}}closest\\ words\end{tabular} & \begin{tabular}[c]{@{}c@{}}harshly, repeatedly, \\ insisting, admitting,\\ accusing, behalf\end{tabular} & \begin{tabular}[c]{@{}c@{}}economical, because, \\ therefore, considering,\\ practical, ensure\end{tabular} & \begin{tabular}[c]{@{}c@{}}team, league, win,\\ championship, scoring,\\ match, final\end{tabular} & \begin{tabular}[c]{@{}c@{}}overseeing, consultant,\\ supervising, executive, \\ deputy, assistant\end{tabular} \\ \hline
example & \begin{tabular}[c]{@{}c@{}}adored him for\\ his personality,\\ despising herself \\ for her eagerness\end{tabular} & \begin{tabular}[c]{@{}c@{}}is costly for \\ the firms,\\ is fantastic for\\ a little boy\end{tabular} & \begin{tabular}[c]{@{}c@{}}desire for friendship,\\ eager for challenge,\\ urge for food\end{tabular} & \begin{tabular}[c]{@{}c@{}}a good chief\\ for the clan,\\ work for a\\ company\end{tabular} \\ \hline
\begin{tabular}[c]{@{}c@{}}TPP\\ sense\end{tabular} & Cause & Experiencer & Beneficiary & Employer \\ \hline
\begin{tabular}[c]{@{}c@{}}sense\\ number\end{tabular} & 8 & 9 & 10 & 11 \\ \hline
\begin{tabular}[c]{@{}c@{}}closest\\ words\end{tabular} & \begin{tabular}[c]{@{}c@{}}outstanding, best, award,\\ exemplary, achievement,\\ recognizing, exceptional\end{tabular} & \begin{tabular}[c]{@{}c@{}}during, spent, \\ beforehand, \\ before, after\end{tabular} & \begin{tabular}[c]{@{}c@{}}fees, expenses, payment,\\ dues, money, pay,\\ taxes, allowance, subsidy\end{tabular} & \begin{tabular}[c]{@{}c@{}}facilitate, assit, enable,\\ using, simplify, simulate,\\ ensure, optimize, purpose\end{tabular} \\ \hline
example & \begin{tabular}[c]{@{}c@{}}outstanding for cuisine,\\ proclivity for risk\end{tabular} & \begin{tabular}[c]{@{}c@{}}for the end of this year,\\ forecast for 1993\end{tabular} & \begin{tabular}[c]{@{}c@{}}higher prices \\ for goods,\\ charge drives for \\ emission tests\end{tabular} & \begin{tabular}[c]{@{}c@{}}defer the issue\\ for later discussion,\\ excellent for \\ this purpose\end{tabular} \\ \hline
TPP sense & Concomitant & TimePeriod & SwapGoal & Purpose \\ \hline
\begin{tabular}[c]{@{}c@{}}sense\\ number\end{tabular} & 12 &  &  &  \\ \hline
\begin{tabular}[c]{@{}c@{}}closest \\ words\end{tabular} & \begin{tabular}[c]{@{}c@{}}hence, because, ample,\\ adequate, consequently\end{tabular} &  &  &  \\ \hline
example & \begin{tabular}[c]{@{}c@{}}suffice for a \\ murder conviction,\\ adequate for flow\end{tabular} &  &  &  \\ \hline 
TPP sense & ReferentNorm &  &  &  \\ \hline
\end{tabular}}
\caption{Senses of Preposition \emph{for}}
\label{tab:for_senses}
\end{table}

Table~\ref{tab:for_senses} provides \emph{for}'s senses.  First look at the synonyms among the nearest neighbors. Word \emph{during} corresponds to the sense TimePeriod, and  ``forecast \emph{for} 1993'' can be replaced with ``forecast \emph{during} 1993''. Then we can find the governors and complements as neighbors. For example, \emph{buy} is a governor in phrase ``buy it for \$200'', and also close to the sense Price. Word \emph{definitions} governs the preposition in phrase ``definition for the term'' and stay close to the sense Referent. Word \emph{outstanding} is also a governor in ``outstanding for a cuisine'', and close to \emph{for}'s sense  Concomitant.
As for complements, \emph{purpose} acts as an complement in ``excellent for this purpose'', and appears close to \emph{for}'s sense of Purpose.

\begin{table}[htbp!]
\centering
\resizebox{1.0\textwidth}{!}{
\begin{tabular}{|c|c|c|c|c|}
\hline
\begin{tabular}[c]{@{}c@{}}sense \\ number\end{tabular} & 0 & 1 & 2 & 3 \\ \hline
\begin{tabular}[c]{@{}c@{}}closest\\ words\end{tabular} & \begin{tabular}[c]{@{}c@{}}pair, featuring,\\ twisted, assorted\end{tabular} & \begin{tabular}[c]{@{}c@{}}using, stacked, \\ resembling, molded,\\ mechanically, adding\end{tabular} & \begin{tabular}[c]{@{}c@{}}signed, contract, \\ professional, manager,\\ career, full-time\end{tabular} & \begin{tabular}[c]{@{}c@{}}switches, microphones,\\ setup, installing,\\ radios, audio\end{tabular} \\ \hline
example & \begin{tabular}[c]{@{}c@{}}rubble with bare hands,\\ nudged Graham \\ with her elbow\end{tabular} & \begin{tabular}[c]{@{}c@{}}healed them with \\ our doctor 's hand, \\ treatment with laser\end{tabular} & \begin{tabular}[c]{@{}c@{}}stint with Somerset,\\ manager with \\ The Northern Echo\end{tabular} & \begin{tabular}[c]{@{}c@{}}a wooden cart \\ with small wheels,\\ the envelope with \\ her resignation\end{tabular} \\ \hline
TPP sense & MeansName & MeansName & Employer & Accountrement \\ \hline
\begin{tabular}[c]{@{}c@{}}sense\\ number\end{tabular} & 4 & 5 & 6 & 7 \\ \hline
\begin{tabular}[c]{@{}c@{}}closest \\ words\end{tabular} & \begin{tabular}[c]{@{}c@{}}scholar, studies, \\ professor, doctoral\end{tabular} & \begin{tabular}[c]{@{}c@{}}treasurer, leader\\ elected, deputy\end{tabular} & \begin{tabular}[c]{@{}c@{}}while, alongside,\\ mutual, befriend, \\ interpersonal\end{tabular} & \begin{tabular}[c]{@{}c@{}}community, \\ voluntary, facilitate,\\ implementing\end{tabular} \\ \hline
example & \begin{tabular}[c]{@{}c@{}}studies literature with \\ the Open University\end{tabular} & \begin{tabular}[c]{@{}c@{}}the value of benefits\\ rises with income,\\ fantastic with the day\end{tabular} & \begin{tabular}[c]{@{}c@{}}partner with \\ systems integrators,\\ conspire with enemy\end{tabular} & \begin{tabular}[c]{@{}c@{}}complied with \\ their obligation,\\ conform with \\ the legislation\end{tabular} \\ \hline
TPP sense & Partner & Coresultant & Accompanier & Harmonizer \\ \hline
\begin{tabular}[c]{@{}c@{}}sense\\ number\end{tabular} & 8 & 9 & 10 & 11 \\ \hline
\begin{tabular}[c]{@{}c@{}}closest\\ words\end{tabular} & \begin{tabular}[c]{@{}c@{}}express, emotional,\\ jealousy, fearful\end{tabular} & \begin{tabular}[c]{@{}c@{}}against, teammate,\\ throwing, punching\end{tabular} & \begin{tabular}[c]{@{}c@{}}news, reporter, press,\\ interviewing, \\ announcing\end{tabular} & \begin{tabular}[c]{@{}c@{}}collar, wears, shoulders,\\ waist, belly, neck\end{tabular} \\ \hline
example & \begin{tabular}[c]{@{}c@{}}glistened with dew,\\ shimmers with crystals\end{tabular} & \begin{tabular}[c]{@{}c@{}}the showdown \\ with his father,\\ collided with \\ a bus\end{tabular} & \begin{tabular}[c]{@{}c@{}}contact me\\ with ideas,\\ call me with \\ an arrangement\end{tabular} & \begin{tabular}[c]{@{}c@{}}people with \\ disabilities,\\ a lady with \\ a pale face\end{tabular} \\ \hline
TPP sense & FeatureCause & Opponent & Message & Attribute \\ \hline
\begin{tabular}[c]{@{}c@{}}sense\\ number\end{tabular} & 12 & 13 & 14 &  \\ \hline
\begin{tabular}[c]{@{}c@{}}closest\\ words\end{tabular} & \begin{tabular}[c]{@{}c@{}}mutual, mutually,\\ resulting, when, \\ thus, meanwhile\end{tabular} & \begin{tabular}[c]{@{}c@{}}symptoms, prognosis, \\ syndrome, abnormalities\end{tabular} & \begin{tabular}[c]{@{}c@{}}dazed, furiously, \\ relentlessly, taunt\end{tabular} &  \\ \hline
example & \begin{tabular}[c]{@{}c@{}}reason with her,\\ compatible with \\ autonomy\end{tabular} & \begin{tabular}[c]{@{}c@{}}woke with a \\ heavy head,\\ awoke with a start\end{tabular} & \begin{tabular}[c]{@{}c@{}}forecast with certainty,\\ express it with \\ passionate intensity\end{tabular} &  \\ \hline
TPP sense & Concomitant & Malady & Manner\&Mood &  \\ \hline
\end{tabular}}
\caption{Senses of Preposition \emph{with}}
\label{tab:with_senses_appendix}
\end{table}

The senses of \emph{with} are listed in Table~\ref{tab:with_senses_appendix}.  A nearest neighbor of sense MeansName is \emph{using}, and sentence ``treatment with laser'' can be rewritten as ``treatment using laser''. A nearest neighbor of sense Accompanier is \emph{alongside}, and ``partner \emph{with} systems integrators'' can be understood as ``partner \emph{alongside} systems integrators''.  \emph{Against} is synonymous to the sense Opponent, and ``collided \emph{with} a bus'' can be paraphrased as ``collided \emph{against} a bus''.
Besides semantically similar words, governors and complements can be included as nearest words.
\emph{Abnormalities} is a governor in ``woke with abnormalities'', and is one neighbor of sense Malady. \emph{News} is a complement in ``contact me with the recent news'', and close to \emph{of}'s sense Message.

These tables of sense representations show us that preposition sense-specific embedding carries non-trivial lexical semantics. Nearest neighbors give a qualitative evaluation of these representations in terms of word similarity. We do observe semantically-similar words are included as nearest neighbors, which can be treated as definitions of the specific preposition sense. We also find that these nearest words reveal the attachment and complement properties of prepositions. Governors and complements may appear close to the given sense.

\subsection{Preposition senses as relations}

\begin{figure*}[htbp!]
\centering
\begin{minipage}{0.32\textwidth}
\centerline{\includegraphics[width=\linewidth]{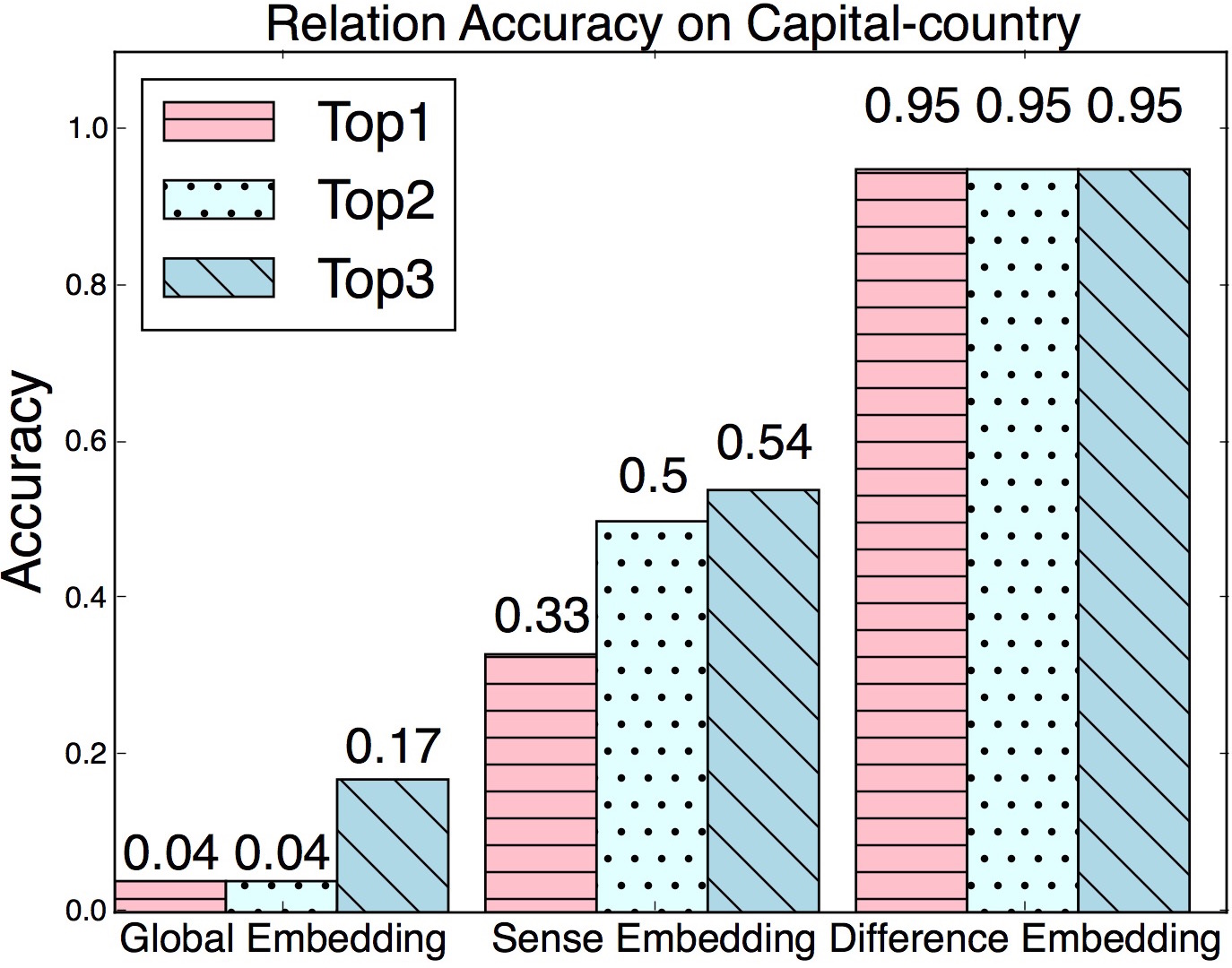}}
\centerline{\small{(a) `from' for capital-country relation}}
\label{fig:from_capital_country}
\end{minipage}
\begin{minipage}[c]{0.32\textwidth}
\centerline{\includegraphics[width=\linewidth]{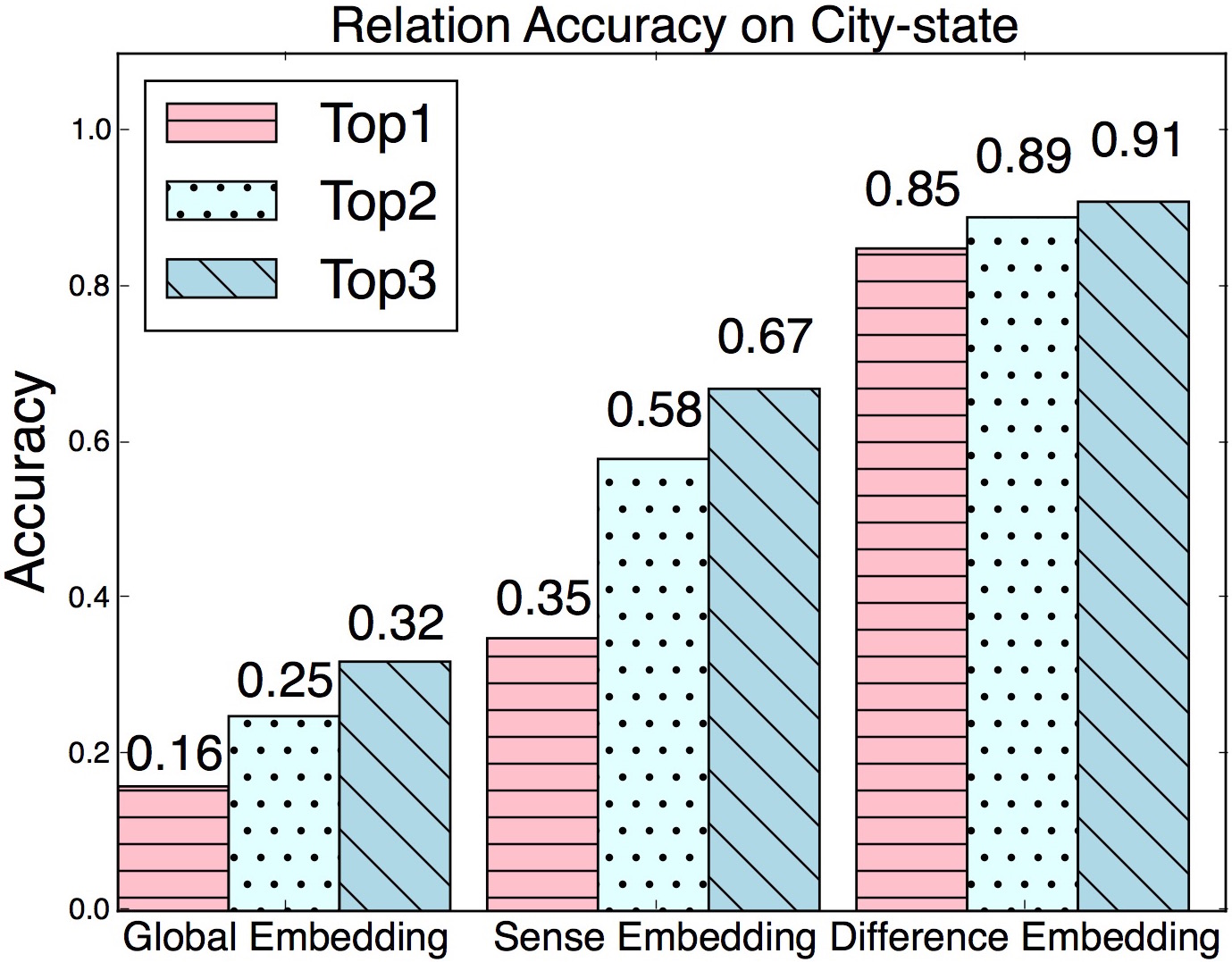}}
\centerline{\small{(b) `in' for city-state relation}}
\label{fig:in_city_state}
\end{minipage}
\begin{minipage}{0.32\textwidth}
\centerline{\includegraphics[width=\linewidth]{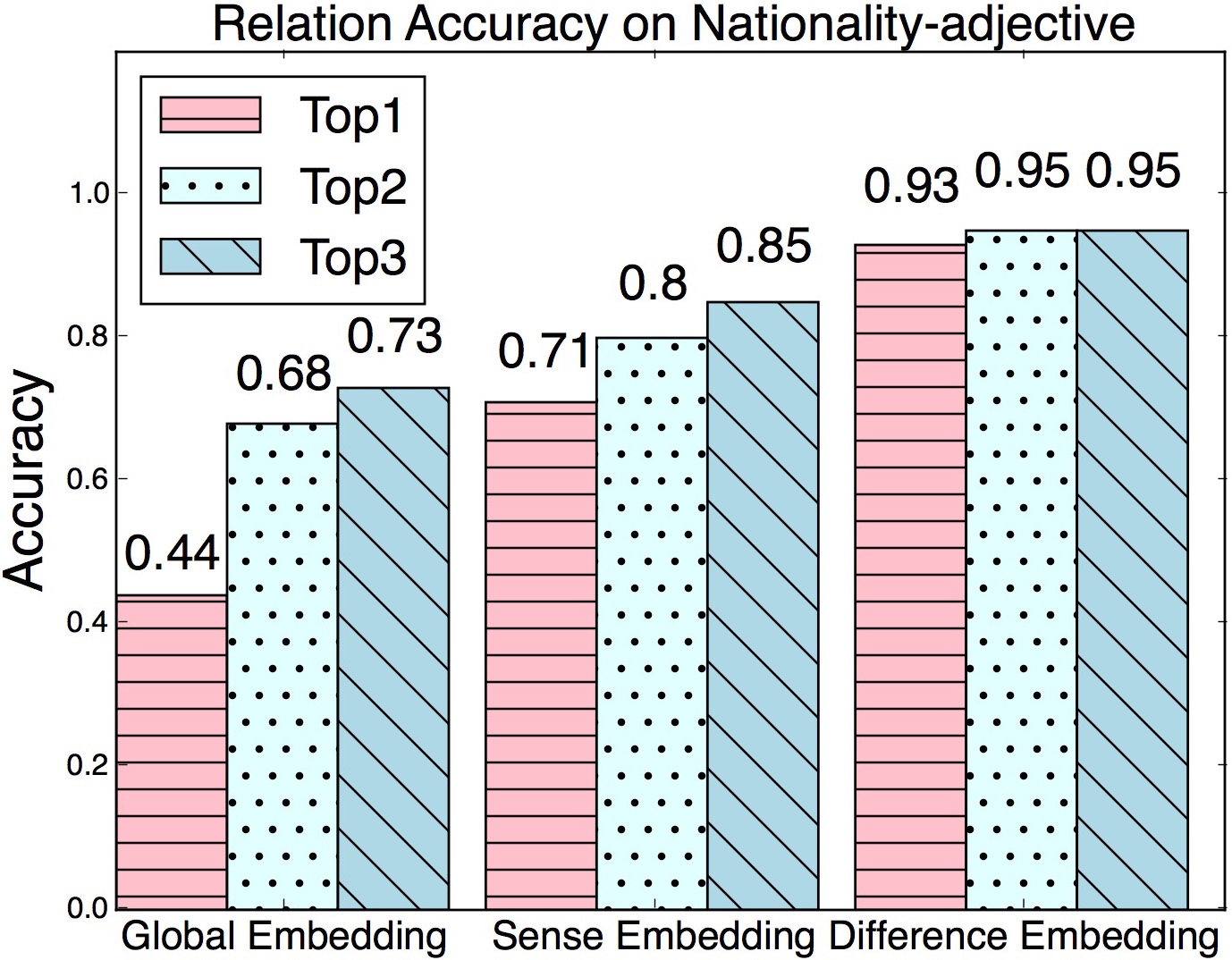}} 
\centerline{\small{(c) `in' for country-nationality relation}}
\label{fig:nationality-adjective}
\end{minipage}
\caption{Sense specific preposition embeddings serve as good approximations of three semantic relations. }
\label{fig:prep_relation}
\end{figure*}

In this section, we use preposition sense to model lexical relations, and predict one word (e.g., country) from the other (e.g., capital). Three candidate predictions are generated from the approximate embedding. In Fig.~\ref{fig:prep_relation}, we report the accuracy of finding the target word (country, state or adjective) in the top $k (k=1,2,3)$ neighbors corresponding to the use of the global embedding, the sense-specific embedding of (`in' and `from') and the difference embedding.

\subsection{Preposition senses aid paraphrasing}
\label{app:paraphrase}

\begin{table*}[htbp!]
\centering
\resizebox{1.0\textwidth}{!}{
\begin{tabular}{|c|c|c|c|c|}
\hline
\multirow{2}{*}{sentence} & \multirow{2}{*}{phrasal verb} & \multicolumn{3}{c|}{paraphrasing} \\ \cline{3-5} 
 &  & \begin{tabular}[c]{@{}c@{}}sense \end{tabular} & \begin{tabular}[c]{@{}c@{}}global \end{tabular} & \begin{tabular}[c]{@{}c@{}}simplex \end{tabular} \\ \hline
\begin{tabular}[c]{@{}c@{}}She could not \textbf{keep from} crying,  and agitated on the chair.\end{tabular} & keep from & \textbf{avoid} & get & maintain \\ \hline
Without a word he leaned forward and switched on the engine. & switched on & \textbf{starting} & shifted & reverted \\ \hline
I have certainly been \textbf{kicked in} the teeth by those bastards. & kicked in & \textbf{knocked}  & throw & \textbf{knocked} \\ \hline
I have chosen to \textbf{block off} the easy track and so turn it into a dead end. & block off & \textbf{stopped} & cleared & cleared \\ \hline
The Rishon Le Zion killings \textbf{sparked off} a wave of sympathy protests. & sparked off & \textbf{ensued} & \textbf{spurred} & \textbf{ignited} \\ \hline
Stanley \textbf{put down} his paper and glared at her. & put down & \textbf{laid} & slammed & brought \\ \hline
\end{tabular}}
\caption{Paraphrasing of Phrasal Verbs}
\label{tab:vp_paraphrasing_appendix}
\end{table*}

In the experiment on phrasal verb paraphrasing, we use preposition global embedding, simplex embedding and our sense-specific  preposition  embedding to approximate the representation of phrasal verbs. The nearest verbs of the phrasal representation are used (excluding the verb in the phrase) as its paraphrases. Some examples of phrasal verbs and paraphrases are shown in Table~\ref{tab:vp_paraphrasing_appendix}, and valid paraphrases are highlighted.
 
For each approximate phrasal embedding ($v_{\text{vp}}^{\text{sense}}$, $v_{\text{vp}}^{\text{global}}$,$v_{\text{vp}}^{\text{simplex}}$), we list the 
nearest three verbs (excluding the verb in the phrase) as candidate paraphrases. Here, the distance is measured in terms of the cosine similarity between the word vectors. 

\begin{figure}[htbp!]
\centering
\includegraphics[width=0.7\textwidth]{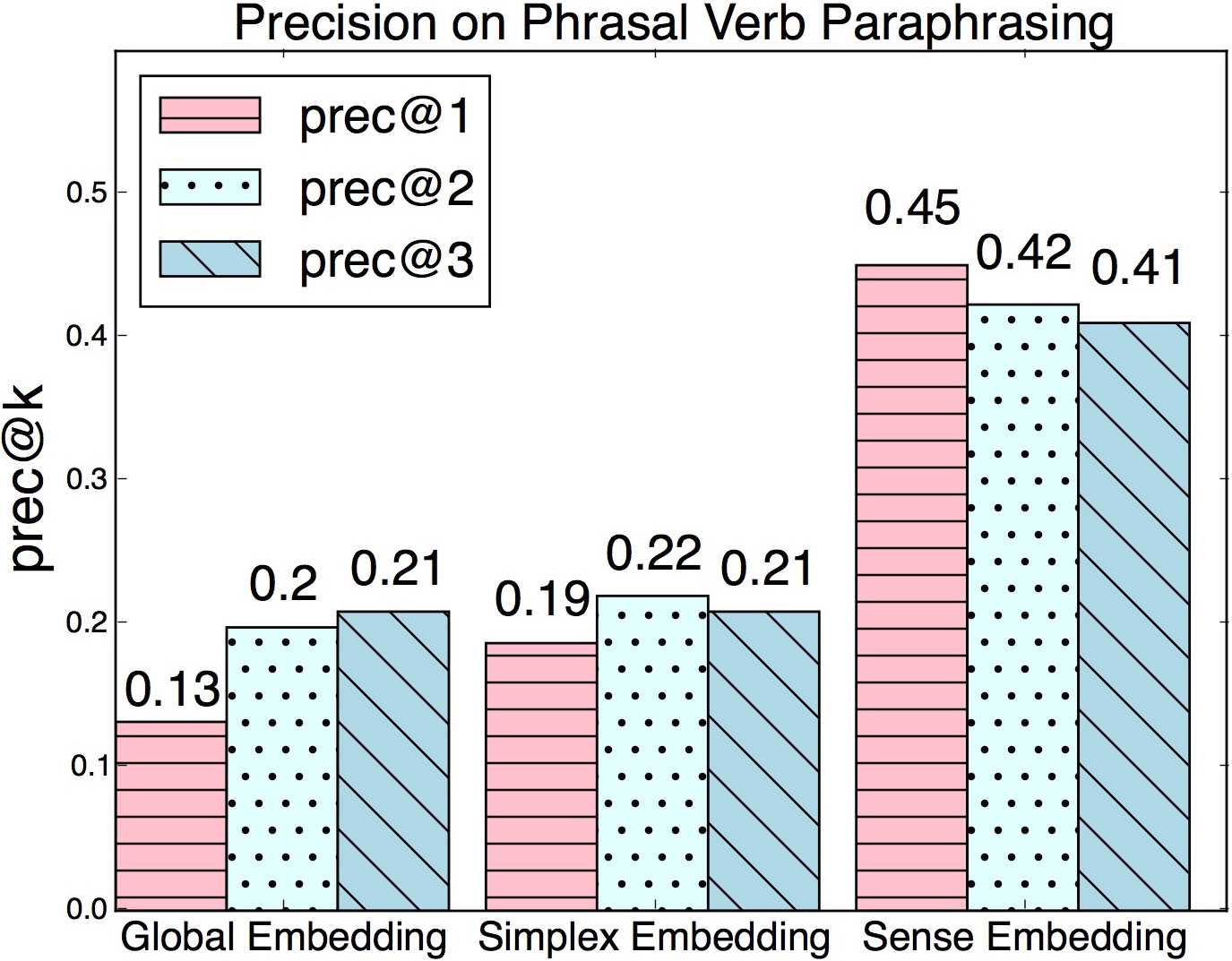}
\caption{Precision on Phrasal Verbs Paraphrasing}
\label{fig:vp_paraphrase}
\end{figure}

Since we listed the top three candidate paraphrases for a phrasal verb and consider the validity, we choose metric precision at $k$ (prec@k) which is defined as: $$\text{prec@k}=\frac{1}{N}\sum\limits_{i=1}^{N}\text{Precision}(i, k),$$  where $N$ is the number of phrasal verbs, and $\text{Precision}(i, k)$ is the percent of good paraphrases among the top $k$ paraphrases for phrase $i$. The precision metrics for each method are reported in Fig.~\ref{fig:vp_paraphrase}. As we can see, our sense-specific preposition  embedding has a significantly better performance than global and simplex embeddings,  in terms of all the three prec@1, prec@2 and prec@3 metrics.

We notice that  paraphrasing is closely related with the nature of phrasal verbs. A three way classification is adopted in \cite{dehe2002particle,jackendoff2002english,emonds1985unified,villavicencio2006verb}, where  verb particle compounds (VPC) can be classified into compositional, idiomatic or aspectual. For the compositional VPCs, the meaning of the construction is determined by the literal interpretations of the particle and the verb (e.g., throw out). Idiomatic VPCs, however, cannot have their meaning determined by their component words (e.g., \emph{get through} meaning `manage to deal with'). The third class, aspectual VPCs, have the particle providing the verb with an endpoint, describing the action in more details (e.g.,  tear up). 


The dataset of English phrasal verbs consists of $91$ phrases, in which there are 54 compositional phrases, 16 aspectual phrases and 21 noncompositional phrases in our dataset. Here we report the precision@k (k=1,2,3) of different methods on these three types of verb phrases respectively.

As is shown in Table~\ref{tab:vp_para_detail}, the precision of paraphrasing with  preposition sense embedding is higher than baselines with preposition global embedding and verb embedding on three types of phrases. As we can see from the table, the precision improvement of sense embedding over global embedding is larger on aspectual phrases than on compositional phrases. The reason might be that preposition plays a more important semantic role in aspectual phrases than in compositional phrases.

We also observe that the precision achieved by simplex embedding is close to precision by global embedding. It means the phrasal verb representations with and without global embedding do not differ too much, which indicates that global embedding does not provide necessary semantic information of prepositions in paraphrasing phrasal verbs.

Also, we find that paraphrasing of compositional or aspectual phrasal verbs is better than that of idiomatic ones. This is because component words do not give much information about the semantics of idiomatic  phrases. Hence addition of components is not a good approximation of idiomatic phrasal representation.

Empirically, phrasal approximation using addition of verb and particle gives good paraphrasing mainly in the following cases:
\begin{enumerate}
\item verb dominates the phrasal meaning, e.g., focus on ($\sim$ focus), carry in ($\sim$ carry);
\item preposition dominates the phrasal meaning, e.g., go against ($\sim$ against), keep from ($\sim$from, one sense of 'from' is close to 'stop' and 'prevent');
\item verb is polysemous, and preposition helps disambiguate the verb. For example, "headed down" where the verb "headed" have two senses: "chaired/led" and "approached". The phrase "headed down" prompts that "headed" should have the sense "approached".
\end{enumerate}

\begin{table}[]
\centering
\begin{tabular}{|c|c|c|c|c|c|c|c|c|c|}
\hline
phrase type & \multicolumn{3}{c|}{Compositional} & \multicolumn{3}{c|}{Aspectual} & \multicolumn{3}{c|}{Idiomatic} \\ \hline
embedding & global & simplex & sense & global & simplex & sense & global & simplex & sense \\ \hline
prec@1 & 0.125 & 0.232 & 0.482 & 0.0625 & 0.0625 & 0.5 & 0.190 & 0.143 & 0.333 \\ \hline
prec@2 & 0.196 & 0.277 & 0.429 & 0.125 & 0.0625 & 0.406 & 0.238 & 0.167 & 0.429 \\ \hline
prec@3 & 0.220 & 0.268 & 0.417 & 0.188 & 0.042 & 0.438 & 0.190 & 0.159 & 0.381 \\ \hline
\end{tabular}
\caption{Precision on Verb Phrase Paraphrasing}
\label{tab:vp_para_detail}
\end{table}

\end{document}